\documentclass[letterpaper, 10 pt, conference]{ieeeconf}  % Comment this line out if you need a4paper

\IEEEoverridecommandlockouts                              % This command is only needed if 
\newcommand{\q}{\boldsymbol{q}}
\newcommand{\qd}{\dot{\boldsymbol{q}}}
\newcommand{\qm}{\boldsymbol{q}_{{m}}}
\newcommand{\qb}{\boldsymbol{q}_{{b}}}

\newcommand{\qca}{\boldsymbol{q}_{ca}}

\usepackage{epsfig} % for postscript graphics files
\usepackage{amsmath} % assumes amsmath package installed
\usepackage{amssymb}  % assumes amsmath package installed
\usepackage{graphicx}

\usepackage[font=small]{caption}
\usepackage{multirow}     % 单元格合并
\usepackage{float}
\usepackage{bm}  % 加粗
\usepackage{cite}  % 连续引注用破折号
\usepackage{hyperref} %添加超链接
\hypersetup{
    hidelinks, % 隐藏所有链接的颜色和边框
}

\title{\LARGE \bf
CushionCatch: A Compliant Catching Mechanism for Mobile Manipulators via Combined Optimization and Learning
}

\author{    Bingjie Chen$^{\dagger}$, Keyu Fan$^{\dagger}$, Qi Yang, Yi Cheng, Houde Liu, \\
    \hspace{1cm} Kangkang Dong, Chongkun Xia, Liang Han, Bin Liang
    \thanks{This work was supported by National Natural Science Foundation of China (No.62203260, 92248304), The Shenzhen Science Fund for Distinguished Young Scholars (RCJC20210706091946001), Tsinghua SIGS Cross Research and Innovation Fund (JC2021005).  ({$\dagger$} indicates equal contribution) (Corresponding authors: Houde Liu, Chongkun Xia (liu.hd@sz.tsinghua.edu.cn, xiachk5@mail.sysu.edu.cn))  }
    \thanks{ Bingjie Chen, Keyu Fan, Qi Yang, Yi Cheng and Houde Liu are with the Center for Artificial Intelligence and Robotics, Shenzhen International Graduate School, Tsinghua University, Shenzhen 518055, China.}
    \thanks{Kangkang Dong is with the Jianghuai Advanced Technology Center, Hefei, 230000, China; and with University of Science and Technology of China, Hefei, 23000, China.}
    \thanks{Chongkun Xia is with the School of Advanced Manufacturing, Sun Yat-Sen University, Shenzhen 518055, China.}
    \thanks{Liang Han is with the School of Electrical and Automation Engineering, Hefei University of Technology, Hefei 230009, China.}
    \thanks{Bin Liang is with the Navigation and Control Research Center, Department of Automation, Tsinghua University, Beijing 100084, China}
    \thanks{$^1{\text{Code}}$:\href{https://github.com/bingjiechen-THU/CushionCatch}{https://github.com/bingjiechen-THU/CushionCatch}}
}

\begin{document}

\maketitle
\thispagestyle{empty}
\pagestyle{empty}

%%%%%%%%%%%%%%%%%%%%%%%%%%%%%%%%%%%%%%%%%%%%%%%%%%%%%%%%%%%%%%%%%%%%%%%%%%%%%%%%
\begin{abstract}
Catching flying objects with a cushioning process is a skill commonly performed by humans, yet it remains a significant challenge for robots. In this paper, we present a framework that combines optimization and learning to achieve compliant catching on mobile manipulators (CCMM). First, we propose a high-level capture planner for mobile manipulators (MM) that calculates the optimal capture point and joint configuration. Next, the pre-catching (PRC) planner ensures the robot reaches the target joint configuration as quickly as possible. To learn compliant catching strategies, we propose a network that leverages the strengths of LSTM for capturing temporal dependencies and positional encoding for spatial context (P-LSTM). This network is designed to effectively learn compliant strategies from human demonstrations. Following this, the post-catching (POC) planner tracks the compliant sequence output by the P-LSTM while avoiding potential collisions due to structural differences between humans and robots. We validate the CCMM framework through both simulated and real-world ball-catching scenarios, achieving a success rate of 98.70\% in simulation, 92.59\% in real-world tests, and a 28.7\% reduction in impact torques. The open source code has be released for the reference of the community$^1$.

\end{abstract}

%%%%%%%%%%%%%%%%%%%%%%%%%%%%%%%%%%%%%%%%%%%%%%%%%%%%%%%%%%%%%%%%%%%%%%%%%%%%%%%%
\section{INTRODUCTION}
Manipulating dynamic objects is challenging, requiring a sophisticated interplay of object detection, motion planning, and control \cite{5651175}. In dynamic manipulation scenarios, tasks such as throwing \cite{Liu2022a, 9560866}, juggling \cite{Woodruff2023}, and catching \cite{kim2014catching, salehian2016dynamical} are essential. Catching dynamic objects requires quick and precise motion planning, followed by fast and accurate execution. For the motion planning of catching, the process can be divided into two phases, pre-catching (PRC) and post-catching (POC) \cite{Zhao2023b}. In the POC phase, for a high-stiffness robot, the object may bounce away or generate significant impact forces upon contact. To address these problems, incorporating a cushioning mechanism during catching is beneficial, as it reduces impact forces \cite{9691802, Haddadin2009}, prevents damage to both the robot and the object \cite{6343846}, and minimizes the risk of rebound, especially for high-speed or high-mass objects \cite{9636775, 9341246}. Moreover, since humans always exhibit cushioning behavior when catching high-mass objects, an effective approach would be for the robot to learn this motion from humans \cite{Zhao2023b}.

\begin{figure}[t]
	\centering
	\includegraphics[width= 0.4 \textwidth,height=5.5cm]{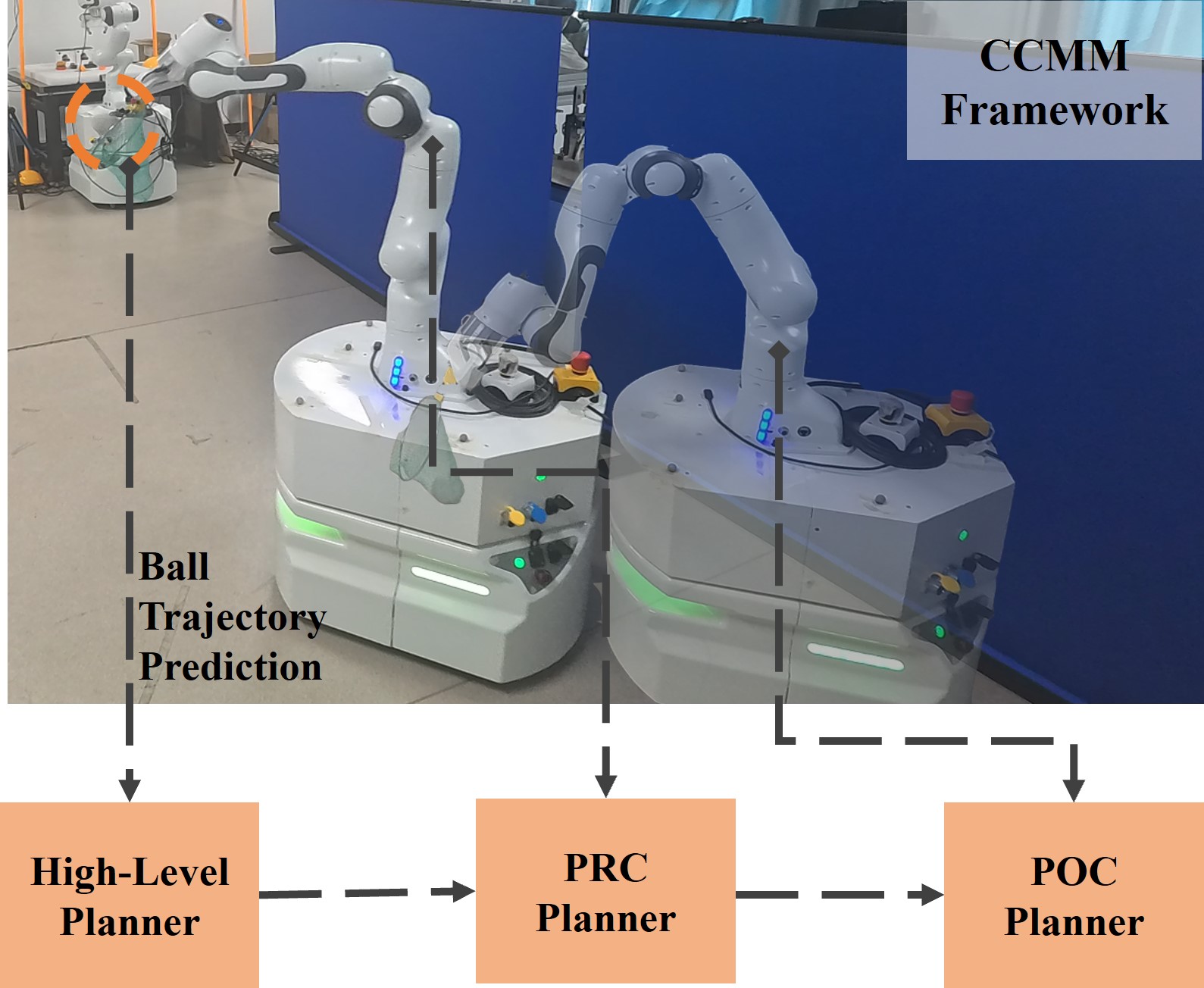}
	\caption{The mobile manipulator's motions under CCMM framework for achieving compliant catching.}
\label{general}
\end{figure}

Mobile manipulators, with their high degrees of freedom, are better suited to learning from human behavior, enabling them to outperform conventional stationary robots in terms of flexibility and efficiency \cite{zeng2020tossingbot, Lampariello2011, Wang2023, Mashali2018, Xie2024}. Their mobility allows them to actively navigate and interact with objects \cite{rizzi2023robust, pankert2020perceptive}. In catching tasks, most prior research focused on rigid control approaches, often overlooking the importance of compliant control \cite{5651175,salehian2016dynamical, Lampariello2011,Dong2020}. We focus on introducing a framework to achieve compliant catching for mobile manipulators (CCMM). Compared with previous works, the contributions of this letter are:

1. A novel CCMM framework effectively addresses all the planning and tracking challenges throughout the entire compliant catching process for mobile manipulators, while being designed for modularity.

2. Propose a P-LSTM network that combines Long Short-Term Memory (LSTM) with positional encoding (PE), designed to effectively learn compliant catching strategies from human demonstrations. Address the potential collisions arising from the structural differences between humans and robots when learning demonstrations.

3. Simulation and physical experiments were conducted on a mobile manipulator to validate the effectiveness of our framework and the role of compliance. A catching success rates of 98.70\% in simulation, 92.59\% in physical experiments and a 28.7\% reduction in impact torque were achieved.

\section{Related Work}

Many studies have explored ball catching tasks using fixed-base manipulators \cite{5651175, salehian2016dynamical, 7139529, 977211}, with a particular focus on motion planning for the manipulator. Early research in this area relied on heuristic methods to determine the optimal grasping point and generated motion trajectories through interpolation \cite{977211}. However, these approaches often fell short in achieving the most effective capture motion for the robot. A more promising approach involves nonlinear optimization, which helps identify the optimal joint configuration for catching the ball, often in combination with parameterization techniques such as five-order polynomials \cite{5651175}. The approach in \cite{salehian2016dynamical} introduces a control law in dynamical systems, allowing the system to perform high-speed manipulations for catching fast-moving objects. However, these methods have typically been limited to fixed-base manipulators.

The work in \cite{Dong2020, Bauml2011} explores the use of mobile manipulators for catching, considering both the manipulator and the mobile base for coordinated control during capture. However, they overlook the importance of compliance. In reality, a ball thrown from a distance can easily impact a joint, exceeding the torque limit \cite{Yan2024}. Current research on cushioning the impact force during capture has focused primarily on fixed-base manipulators \cite{Stouraitis2020, Uchiyama2012}. These studies rely on impedance controller based approaches, which are challenging to implement in mobile manipulators for generating unified, coordinated control.

Learning from human demonstrations is an effective method for teaching robots to execute fast and reactive motions. Humans can catch moving objects with nonzero velocity, creating smooth  trajectories, even for unexpected objects \cite{Kajikawa1999}. In \cite{kim2014catching}, the robot learns motion from human throwing demonstrations, but it halts immediately when the object makes contact with the end-effector. \cite{Zhao2023b} develops nucleated motion primitives from human demonstration data to generate compliant trajectories for catching.  However, this research focuses only on the capture of vertically falling objects and not account for the potential safety issues arising from structural differences between humans and robots. Transferring humans’ compliant behavior to mobile manipulators for catching flying objects while ensuring safety, remains an unresolved challenge.

\begin{figure*}[t]
	\centering
	\includegraphics[width= 0.85 \textwidth,height=7.2cm]{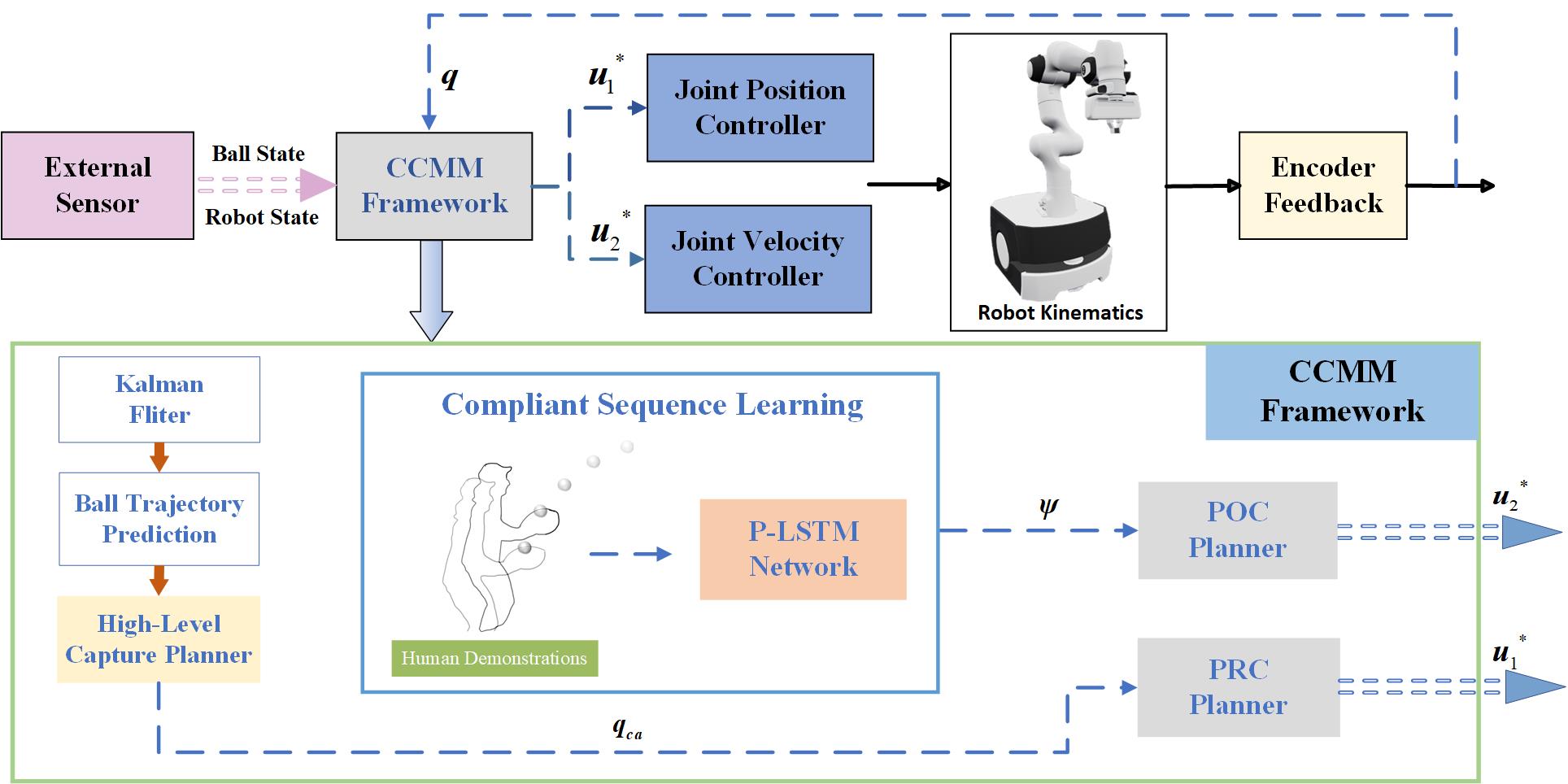}
	\caption{Overview of the system framework: External sensors provide status for both the ball and the robot. The ball’s state is refined using a Kalman filter, which is then used to predict its future trajectory over a time horizon. The high-level capture planner computes the optimal capture joint state, $\q_{ca}$. The PRC planner generates rapid motion trajectories to $\q_{ca}$ and sends them to the joint position controller. The PRC planner tracks the compliant velocity sequence from the P-LSTM network, which is designed to learn from human demonstrations.}
\label{framework}
\end{figure*}

\section{MODELING AND PROBLEM FORMULATION}
\subsection{System Model}
The mobile manipulator we use includes a mobile base and a manipulator, with $n = n_b + n_m$ degrees of freedom. The state of the manipulator is described by joint positions $\qm\in\mathbb{R}^{n_m}$. For a mobile base with a differential drive, we use the full location description $(x_b,y_b,\varphi)$ and define the $\qb \in \mathbb{R}^{n_b}$ as virtual joints. 

For the container fixed at the end for catching the ball, we get the overall forward kinematics as follows:
\begin{equation}
{}_c^w\boldsymbol{T} = {}_b^w\boldsymbol{T}(x_b,y_b,\varphi) \cdot {}_m^b\boldsymbol{T} \cdot {}_e^m\boldsymbol{T}(\qm) \cdot {}_c^e\boldsymbol{T},
\end{equation}
where $w$ represents the world coordinate system; ${}_c^w\boldsymbol{T}$ represents the homogeneous transformation of the container to the world; ${}_b^w\boldsymbol{T}$ is the transformation of the mobile base frame relative to the world; ${}_m^b\boldsymbol{T}$ denotes a constant relative pose from the mobile base frame to the manipulator frame; ${}_e^m\boldsymbol{T}$ is the position forward kinematics of the manipulator where the end-effector (EE) frame is $e$ and ${}_c^e\boldsymbol{T}$ refers to the transformation from the EE frame to the container frame.

Analogous to the Jacobian used in manipulator kinematics, we define the extended Jacobian matrix ${}_c^w\boldsymbol{J}$ for the mobile manipulator to compute the container's velocity in the world coordinate system.
\begin{equation}
    \label{exp_v}
    {}_c^w\boldsymbol{v} = {}_c^w\boldsymbol{J}(x_b,y_b,\varphi ,\qm, {}_c^e\boldsymbol{T})\dot{\boldsymbol{q}},
\end{equation}
where  $\dot{\boldsymbol{q}}=(\dot{\boldsymbol{q}}_b,\dot{\boldsymbol{q}}_m)^\text{T}$ and ${}_c^w\boldsymbol{v}=(v_x,v_y,v_z,\omega_x,\omega_y,\omega_z)^\text{T}$. This maps the velocity of whole-body joints to the container's velocity.

The flying ball is treated as a point mass influenced by aerodynamic drag and gravity \cite{Muller2011}.  The equation governing the ball's motion can be described as:
\begin{equation}
    \label{dy_ball}
    \ddot{\boldsymbol{p}}=-\boldsymbol{g}-k_{ad}\|\dot{\boldsymbol{p}}\|\dot{\boldsymbol{p}},
\end{equation}
where $\boldsymbol{p} \in \mathbb{R}^{3}$ is the ball's position in the world frame, $\boldsymbol{g}=[0, 0, g]^\text{T}$ is the gravitational acceleration, $k_{ad}$ is the aerodynamic drag coefficient and $\|\cdot\|$ is the Euclidean norm.

\section{METHODOLOGY}
\subsection{Ball Estimation and Prediction}
To predict the trajectory of a ball in flight, we should first estimate its current state and then use a dynamical model to forecast its future position over a given time horizon. We use a modified discrete-time Kalman filter that accounts for aerodynamic drag, as described in \cite{Muller2011}. Given measurements at the discrete time index $k$, the state of the ball at the next time step $\boldsymbol{p}[k+1|k]$ can be predicted as follows:
\begin{subequations}
    \begin{gather}
    \ddot{\boldsymbol{p}}[k|k]=-\boldsymbol{g}-k_{ad}||\dot{\boldsymbol{p}}[k|k]||\dot{\boldsymbol{p}}[k|k], \\
    \dot{\boldsymbol{p}}[k+1|k]=\dot{\boldsymbol{p}}[k|k]+\delta_k\ddot{\boldsymbol{p}}[k|k], \\
    \boldsymbol{p}[k+1|k]=\boldsymbol{p}[k|k]+{\delta}_{k}\dot{\boldsymbol{p}}[k|k]+\frac{1}{2}{\delta}_{k}^{2} \ddot{\boldsymbol{p}}[k|k],
\end{gather}
\end{subequations}
where $\delta_k$ is the time interval between $k$ and $k+1$. The Kalman filter we used follows the standard approach described in \cite{Barfoot2024}. 
The aerodynamic drag coefficient $k_{ad}$ is estimated offline using a Recursive Least-Squares estimator \cite{Muller2011}. Then the ball's predicted trajectory  point $\hat{\boldsymbol{p}}$ can be obtained  by (\ref{dy_ball}) through numerically integrating over a finite time horizon. We use a cubic spline interpolation between two-time steps to query the $\hat{\boldsymbol{p}}(t)$ and $\hat{\dot{\boldsymbol{p}}}(t)$ at any time.

\subsection{High-Level Capture Planner}
The high-level planner outputs an optimal catching configuration $\boldsymbol{q}_{ca} \in \mathbb{R}^{n}$ and catch time $t_{ca} \in \mathbb{R}$ that enables the container to intercept the ball’s trajectory at time $t_{ca}$. From current state $\q_0$, catching a moving ball requires the container and ball to be aligned in position and parallel in the direction vector. For this purpose, the following optimization problem is formulated:
% equation内嵌gathered，共用一个编号
\begin{subequations}
    \begin{gather}
    \min_{\qca, t_{ca}}\frac12\left( \boldsymbol{\mathit \Lambda} \| \q_{ca} - \q_0 \| ^2 - \alpha \|t_{ca}\|^2 \right)  \label{high_level_min}\\
    \text{s.t.} \quad  {}_c^w\boldsymbol{P}(\qca)=\hat{\boldsymbol{p}}(t_{ca}),  \label{p_equal}\\
    \frac{{}_c^w\boldsymbol{ z}(\qca)}{|{}_c^w\boldsymbol{ z}(\qca)|} = -\frac{\hat{\dot{\boldsymbol{p}}}(t_{ca})}{|\hat{\dot{\boldsymbol{p}}}(t_{ca})|}, \label{dot_p}\\
    {}_c^w\boldsymbol{T}_z(\qca) \geq \beta  \label{T_z}, \\
    \qca^{min} \leq \qca \leq \qca^{max}.
    \end{gather}
\end{subequations}
The objective function tends to minimize joint movement and extend capture time. $\boldsymbol{\mathit\Lambda}$ and $\alpha$ are the weights associated with each aspect. $\boldsymbol{\mathit\Lambda}$ can be defined as:
\begin{equation}
        \label{lam}
	\boldsymbol{\mathit \Lambda}= \begin{pmatrix}
		\boldsymbol{E}({\mathit \Lambda}_b) & \boldsymbol{0} 	\\
		\boldsymbol{0} & \boldsymbol{E}({\mathit \Lambda}_m) \\
	\end{pmatrix},
\end{equation}
where $\boldsymbol{E}$ represents identity matrix,  ${\mathit \Lambda}_b$, ${\mathit \Lambda}_m$ denotes the cost values associated with the mobile base and the manipulator. Due to the motion control accuracy of the mobile base is generally lower than the manipulator, ${\mathit \Lambda}_b$ tends to set higher than ${\mathit \Lambda}_m$ to encourage more movement of the manipulator.

${}_c^w\boldsymbol{P} \in \mathbb{R}^{3} $ represents the position of the container in the world frame. Equation (\ref{p_equal}) ensures that the position of the container is the same as the ball's position at $\qca$ . ${}_c^w\boldsymbol{z}(\qca) \in  \mathbb{R}^{3}$ indicates the container's z-axis vector expressed in the world frame. Through (\ref{dot_p}), the direction of the container is opposite to the velocity of the ball at $t_{ca}$, that is, parallel.

${}_c^w\boldsymbol{T}_z(\qca) \in  \mathbb{R}$ in (\ref{T_z}) denotes the $Z$-axis height of the container at the capture configuration $\q_{ca}$. Due to the cushioning motions in subsequent POC phase, (\ref{T_z}) is meaningful for the robot to maintain a specific height when catching the ball.

\subsection{PRC Planner}
In the high-level capture planner, the robot's joint configuration $\qca$ for catching the ball is determined. The robot then needs to reach this joint configuration as quickly as possible. Considering the maximum acceleration and speed limits for each joint, we determine the minimum trajectory planning time for each joint, denoted as $t_{prc}$:
\begin{equation}
    t_{prc}=\begin{cases}
    2\sqrt{\frac{\triangle q}{\ddot{q}_{max}}}, &\quad \triangle q<2 q_{acc}    \\
    2\frac{\dot{q}_{max}}{\ddot{q}_{max}}+\frac{\triangle q-\ddot{q}_{max}\left(\frac{\dot{q}_{max}}{\ddot{q}_{max}}\right)^2}{\dot{q}_{max}}, &\quad \triangle q \geq 2 q_{acc}
    \end{cases}     
\end{equation}
where $\triangle q = |q_{ca} - q_0|$,  $q_{acc}=\frac{1}{2} \ddot{q}_{max} \left(\frac{\dot{q}_{max}}{\ddot{q}_{max}}\right)^2$. This section determines the shortest theoretical time for the robot to reach the target configuration $q_{ca}$ if following the limits on maximum acceleration and speed. While such an extreme motion is not used in practice, this simplified method helps us quickly estimate the duration of the joint movement. we can define the time parameter $T_{prc}$ of PRC as follow:
\begin{equation}
    \label{T_prc}
    T_{prc} = \lambda \cdot max(t_{prc}^i) \quad i \in [1,n],
\end{equation}
where $\lambda > 1$ is a proportion parameter to enlarge planning time appropriately. Then, we can get the PRC planner $\phi_{prc}$ in joint space:
\begin{equation}
{\mathit \Phi}_{prc} = \text{Poly5}(t;T_{prc},\q_0, \qca)),
\end{equation}
where \text{Poly5()} represents a Quintic Polynomial Planning that maps the time $t$(ranging from 0 to $T_{prc}$) to the joint position based on the inputs $T_{prc},\q_0, \qca$.

\subsection{Compliant Sequence Learning}
Given the velocity and position states of the object before catching, we employ the P-LSTM network to output the compliant sequences required for POC phase.

Specifically, we concatenate the initial six-dimensional input sequence (comprising position and velocity along the $x$,$y$,$z$-axes) into a variable denoted as $input_1$. This serves as the first input at time step 1. This concatenated vector is then input into a one-step LSTM module, which consists of an LSTM cell followed by a fully connected layer that projects the hidden state to the output dimension. Then, the first output $out_1$ is produced at time step 1 by the one-step LSTM. Subsequently, we concatenate $input_1$ and $out_1$ as the second input $input_2$ at time step 2 and again put it into the same one-step-LSTM module to get the second output flow $out_2$ at time step 2. This process repeats until the maximum sequence length is reached. 

\begin{figure}[t]
\centering
\includegraphics[width=0.5\textwidth]{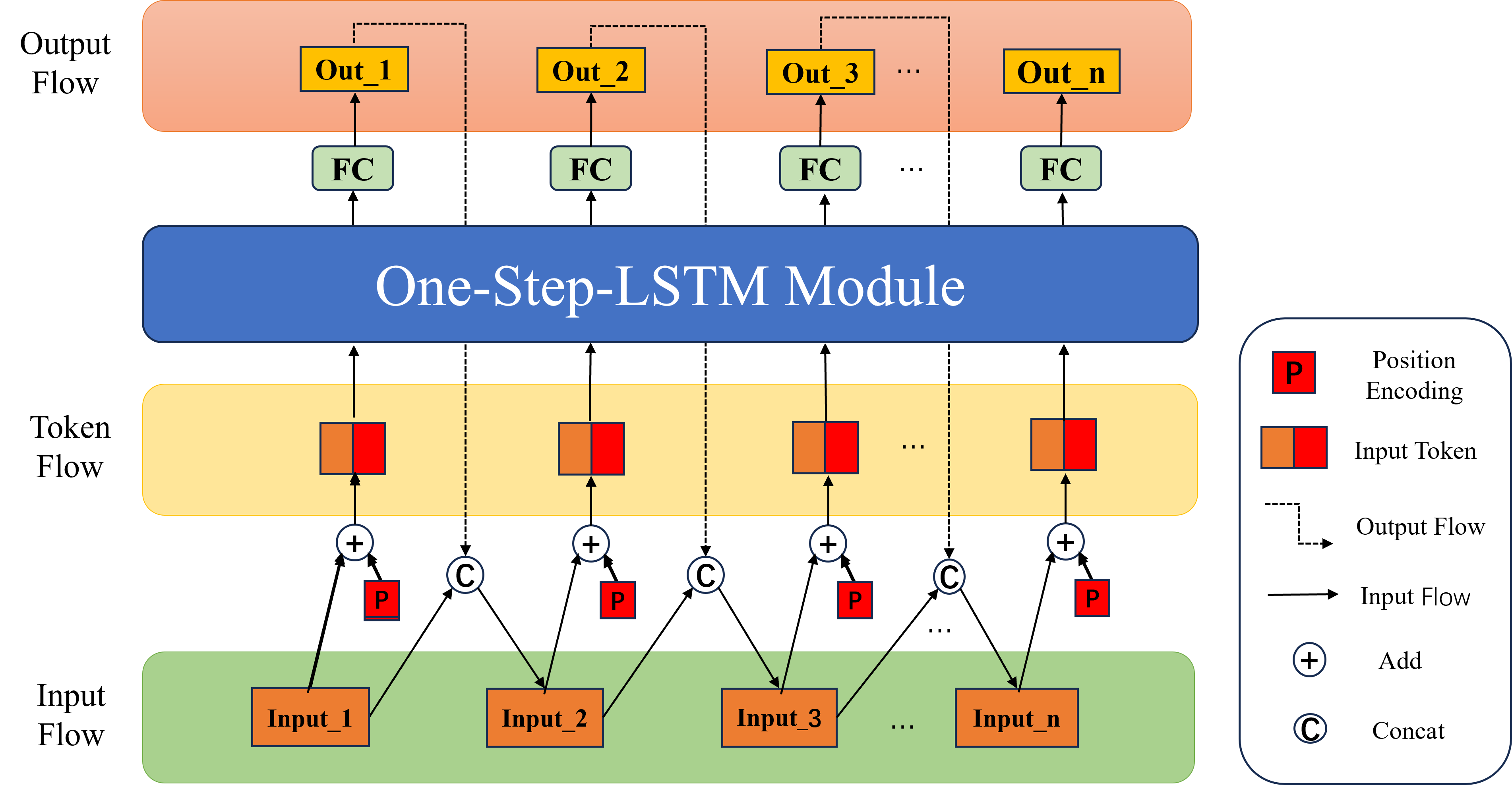} % 图片路径
\caption{P-LSTM network for compliant sequence learning from human demonstrations} % 图片标题
\label{fig:image1} % 图片标签，用于文中引用
\end{figure}

To incorporate the positional order of the input sequences, similar to the approach in \cite{9812087}, we apply position encoding in our LSTM model. Specifically, we encode the sine and cosine of the input position sequence and combine these encodings with the input data. The resulting position-encoded input tokens are then passed into the LSTM model, preserving the order information between sequences for improved learning performance. As shown in Fig. \ref{fig:image1}, the overall output flow can be described as:
\begin{subequations}
    \begin{gather}
        input_{i} = input_{i-1} \oplus out_{i-1}  + PE(i-1 , 6),\\ 
        out_{i} = \mathcal{L}(input_{i}), 
    \end{gather}    
\end{subequations}
where $PE(\cdot)$ means positional encoding and $i$ is the $i$-th time step, $\oplus$ represents concatenation vertically and $\mathcal{L}(\cdot)$ is the one-step-LSTM module.

For each position $l$ in the sequence, the positional encodings are calculated as follows:
\begin{subequations}
    \begin{gather}
        pe(l,2i,k) = \sin\left(\frac{l}{10000^{2i/k}}\right),\\
        pe(l,2i+1,k) = \cos\left(\frac{l}{10000^{2i/k}}\right),\\
        PE(l,k) = {pe(l,0,k)|pe(l,1,k)|...|pe(l,k,k)},
    \end{gather}
\end{subequations}
where $|$ means concatenation horizontally and $k$ is the embedding dimension.

The P-LSTM network is trained on data obtained from human demonstrations. During the application phase, the predicted ball position and velocity at $\qca$ , denoted as  $\hat{{\boldsymbol{p}}}(\qca)$ and $\hat{\dot{\boldsymbol{p}}}(\qca)$, is computed and subsequently used as input to the network. Based on this input, the network generates the compliant velocity sequence $\boldsymbol{\psi} \in \mathbb{R}^{6 \times s}$ , where $s$ denotes the length of the sequence. We focus on tracking the compliant velocity sequence because position is heavily dependent on the world coordinate system. In contrast, velocity is relative and more general, independent of the position of world coordinate system, making it more universal.
\begin{figure}[]
	\centering
	\includegraphics[width= 0.48\textwidth,height=3.3cm]{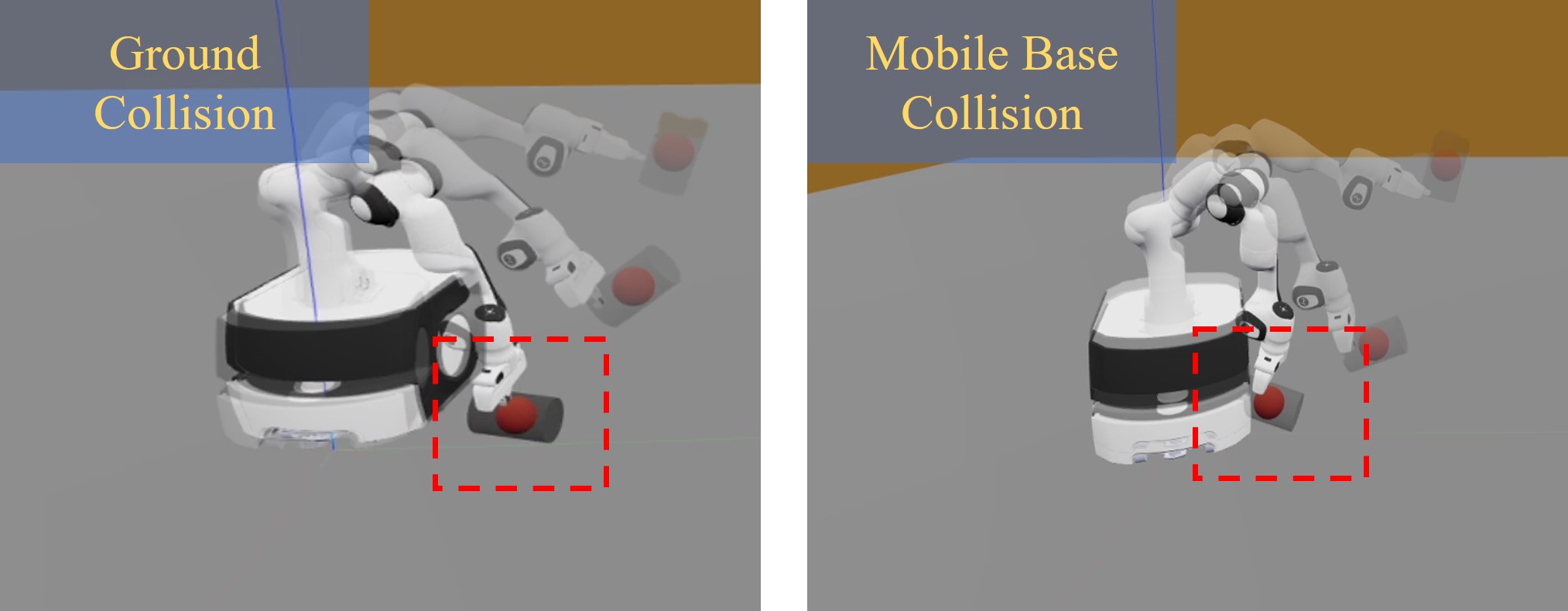}
	\caption{The collision occurs when there are no safety constraints in the POC planner.}
\label{collision}
\end{figure}

\subsection{POC Planner}
After the robot catches the ball, it needs to apply compliant control to have a cushioning mechanism. Given the velocity-compliant sequence $\boldsymbol{\psi}$ produced by the network, we need to track this sequence. Additionally, applying the sequence directly to the robot could result in a series of collisions due to the structural differences between the robot and the human, as shown in Fig. (\ref{collision}). To ensure safety, we must account for potential collisions between the mobile base and the manipulator, as well as between the manipulator and the ground. Therefore, we look at how Control Barrier functions(CBF) \cite{Xu2018} can encode safety constraints. The POC planner can be expressed as follows:
\begin{subequations}
    \begin{gather}
{\mathit \Phi}_{poc} = \min_{\qd}\frac12\left( \| \qd \| ^2 +  \| \boldsymbol{\delta} \|^2 \right )\label{poc_min}\\
    \text{s.t.} \quad  {}_c^w\boldsymbol{J}\qd = \psi_i + \boldsymbol{\delta},  \label{poc_j} \\
    \dot f(z_c) \cdot {}_c^w\boldsymbol{J}_z \cdot \qd \geq -\gamma_z  f(z_c),  \label{poc_z}\\
    \dot g(x_c,y_c) \cdot {}_c^w\boldsymbol{J}_{xy} \cdot \qd \geq - \gamma_{xy}  g(x_c,y_c), \label{poc_xy}
\end{gather}
\end{subequations}
where $\psi_i$ represents the $i$-th sequence of $\boldsymbol{\psi}$ and $\boldsymbol \delta$ is the slack vector to reduce the rigor of the equation. The components of the Jacobian in $z$-axis and the $x,y$-axis are denoted by ${}_c^w\boldsymbol{J}_z$ and ${}_c^w\boldsymbol{J}_{xy}$. The function $f(z_c)$  represents the safe distances between the container and the ground. And $g(x_c,y_c)$ is the safe horizontal distance between the container and the mobile base. Safety is maintained  when $f(z_c)>0$ and $g(x_c,y_c)>0$. In the optimization problem above, equality constraints ensure the container keeps track of expectations $\boldsymbol \psi$. To ensure safety, CBFs are used as inequality constraints, which help maintain safe forward invariance \cite{9561253, 9812378}, that is $f(z_c)>0$ and $g(x_c,y_c)>0$.  By imposing limits on the container’s height along the z-axis and regulating its horizontal distance from the base, the optimization process filters out potential collisions while minimizing deviations from the desired trajectory. 

\begin{figure}[t]
	\centering
	\includegraphics[width= 0.35\textwidth,height=4cm]        {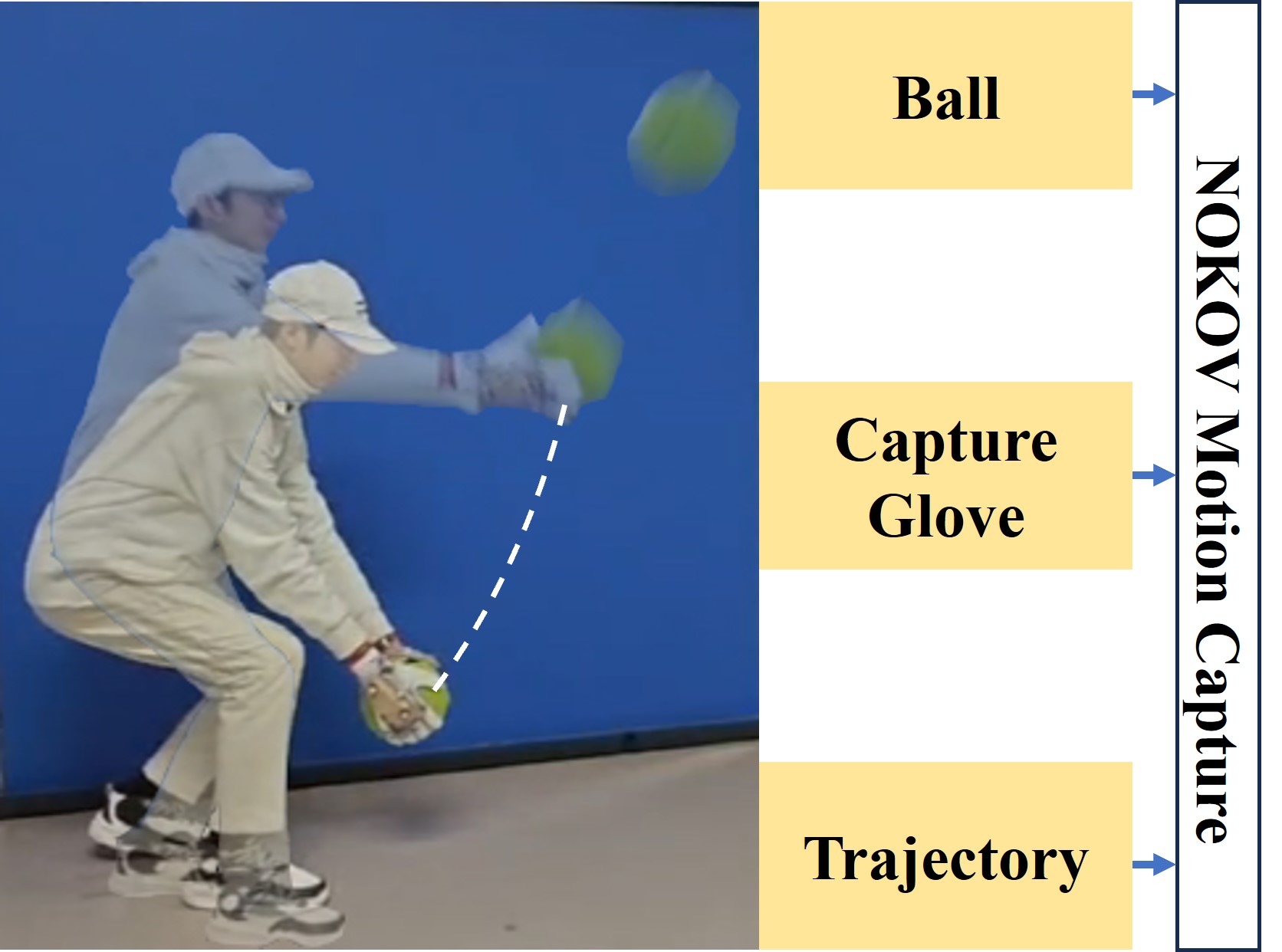}
	\caption{Collecting compliant data from human demonstration in the real world}
\label{collect}
\end{figure}

\begin{figure}[]
	\centering
	\includegraphics[width=0.35\textwidth ,height=4cm]{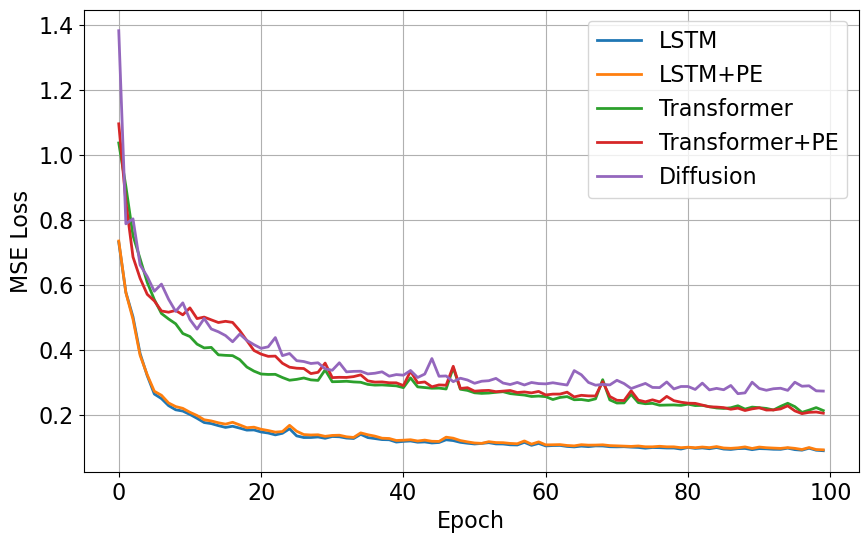}
	\caption{Training loss curves for compliant sequence learning across different networks.   }
\label{loss curve}
\end{figure} 

\begin{figure*}[t]
	\centering
	\includegraphics[width= \textwidth,height=3.3cm]{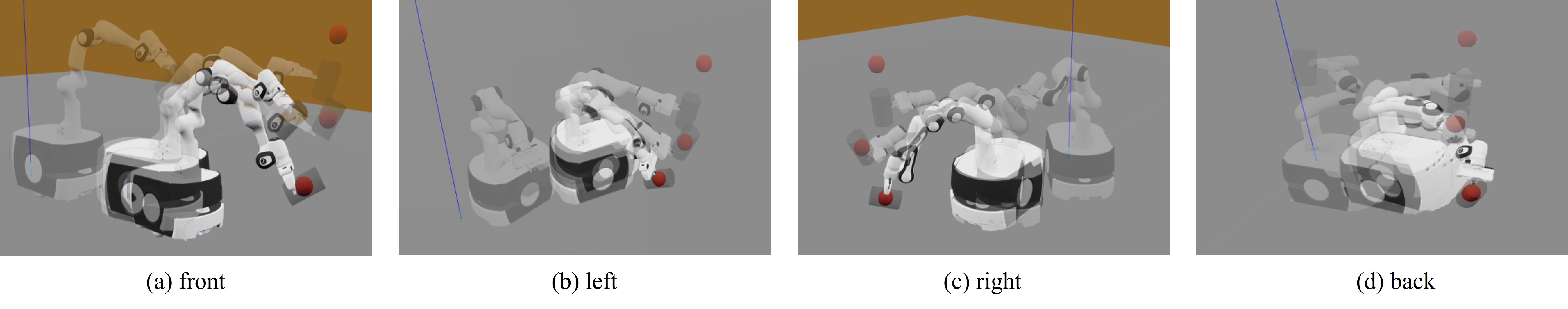}
	\caption{Snapshots from simulation experiments show the ball approaching the robot from various directions, with the robot's opacity gradually increasing over time. }
\label{sim}
\end{figure*}

\begin{figure*}[t]
	\centering
	\includegraphics[width= \textwidth,height=3cm]{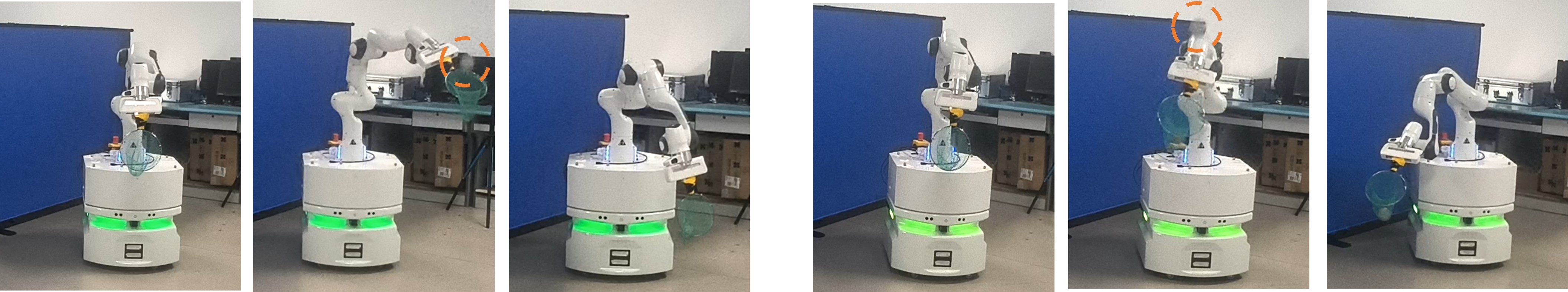}
	\caption{The robot motions under CCMM framework to catch a ball with compliant control. They represent the starting state of motion, the end state of PRC planner, and the end state of POC planner.}
\label{physical}
\end{figure*}

\section{EXPERIMENTS}
We evaluate our framework through experiments conducted both in simulation and on a physical mobile manipulator. We use the following value for the parameters of our framework: $k_{ad}=0.0295$ in (\ref{dy_ball}); $\alpha=2$ in (\ref{high_level_min}); $\beta=0.5$ in (\ref{T_z});  ${\mathit \Lambda}_b = 5, {\mathit \Lambda}_m = 1$ in (\ref{lam}); $\lambda=1.5$ in (\ref{T_prc}); $\gamma_z=0.1$ in (\ref{poc_z}); $\gamma_{xy}=0.1$ in (\ref{poc_xy}).  We use ``SLSQP" method to solve nonlinear optimization problems (\ref{high_level_min}) and (\ref{poc_min}). In P-LSTM, we set the hidden size to 64 and a fully connected layer block from 64 to 6. During training, we use mean squared error (MSE) between the predicted output and the ground truth. The learning rate is set to 0.001, and the max output sequence length is capped at 16. 

\subsection{Data Collection and Processing}

We capture the velocity and position sequences of the ball using a NOKOV motion-capture device during human-catching demonstrations as Fig. (\ref{collect}) shows. Reflective balls are placed on both the glove and the ball. We use the distance between them to determine when to start recording data. In total, 302 trajectories were collected. To distinguish between inputs and outputs of the data for the network, we identified the point with the highest average velocity variation across all data points. This point is then used to split the sequence into input data and the corresponding output labels.

\subsection{Network Comparison Experiment}
To evaluate our P-LSTM method, we compare it with four other methods: LSTM without PE, transformer without PE, transformer with PE, and diffusion policy. All methods use 2 hidden layers with a hidden size of 32, and the transformer's number of heads is set to 2 (the input size is 6, with 3 position components and the remaining 3 being velocities, so 2 heads are sufficient to capture the relationships between them). The training and testing metrics are the MSE between the six-dimensional true sequences and the predicted sequences. We randomly divide the data into a training set (4/5) and a test set (1/5).

As shown in TABLE \ref{ablation} and Fig. \ref{loss curve}, transformer-based methods struggle to converge and validate effectively due to their high parameter demands and the short input sequences. The diffusion policy also fails to train effectively due to the limited data size. Additionally, adding PE to both LSTM and transformer models reduces the error, with almost no impact on inference speed in test experiments.

\subsection{Simulation Scenarios}
We use the Swift and Rtb Python libraries \cite{rtb} as our simulated environments. Initially, we generated 6,000 ball samples $[p_x, p_y, p_z, \dot{p}_x, \dot{p}_y, \dot{p}_z]$, restricting their positions within a cylindrical space of a specified height. We ensured that their trajectories, for $t>0.5$, would pass through the robot's capture range. Then ball parameters are generated randomly by procedure. Additionally, we performed ablation studies on the collision avoidance constraints in (\ref{poc_z}) and (\ref{poc_xy}) during the POC phase (2,000 pieces of data for each test).  The statistical results are presented in TABLE \ref{table_compare1}.

\begin{table}[t]
	\centering
	\caption{Test Experiments}
	\renewcommand{\arraystretch}{1.5} % 设置表格行高为原始行高的1.5倍
	\setlength{\tabcolsep}{6pt} 	  % 设置列之间的间隔为6pt(可以稍微增大间隔)
	\begin{tabular}{ccc}
		\hline
		 Method & Test MSE &  Inference time (s)  \\ 
		\hline
		Transformer & 0.168 & 0.201  \\
        Transformer+PE  & 0.152 & 0.223  \\
        LSTM  & 0.096 & $\textbf{0.061}$  \\
        LSTM+PE & $\textbf{0.091}$ & 0.062  \\
        Diffusion & 0.161 & 0.307  \\
        \hline
	\end{tabular}
	\label{ablation}
\end{table}

\begin{table}[t]
	\centering
	\caption{Simulation results over 6000 random catching tasks}
	\renewcommand{\arraystretch}{1.5} % 设置表格行高为原始行高的1.5倍
	\setlength{\tabcolsep}{4pt} 	  % 设置列之间的间隔为3pt(缩小间隔)
	\begin{tabular}{ccccc}
		\hline
		Approach & Success Rate & Ground Crash &  Base Crash & Not Catch \\ 
		\hline
		General & 98.70\%  & 0.00\%  & 1.05\% &	 0.25\% \\
	Without (\ref{poc_z}) & 82.80\% & 13.70\% & 3.20\%  &0.30\% \\
		Without (\ref{poc_xy}) & 89.40\%  & 0.00\%  & 10.40\% & 0.20\%\\
		\hline
	\end{tabular}
	\label{table_compare1}
\end{table}
Fig. \ref{sim} shows the robot motions of some experiments where the ball comes from different directions. TABLE \ref{table_compare1} demonstrates the results of 6000 random catching tasks, showing that the CCMM framework achieves a 98.70\% success rate for catching. In practice, constraints (\ref{poc_z}) and (\ref{poc_xy}) do not affect the PRC planner, resulting in similar ``Not Catch" rates. However, incorporating collision avoidance constraints notably reduces the incidence of collisions. Without ($\ref{poc_z}$) or ($\ref{poc_xy}$), the probability of collisions in their respective dimensions increases substantially.

\begin{table}[t]
	\centering
	\caption{Process statistics}
	\renewcommand{\arraystretch}{1.5} % 设置表格行高为原始行高的1.5倍
	\setlength{\tabcolsep}{4pt} 	  % 设置列之间的间隔为3pt(缩小间隔)
	\begin{tabular}{ccc}
		\hline
		cap. Error(cm) &  cap. Time(s)  & cal. Time(s) \\ 
		\hline
	    $0.98 \pm 0.44$  & $0.73 \pm 0.12$ &  $0.10 \pm 0.03$ \\
		\hline
	\end{tabular}
	\label{table3}
\end{table}

\begin{table}[t]
	\centering
	\caption{Collision avoidance statistics}
	\renewcommand{\arraystretch}{1.5} % 设置表格行高为原始行高的1.5倍
	\setlength{\tabcolsep}{4pt} 	  % 设置列之间的间隔为3pt(缩小间隔)
	\begin{tabular}{ccc}
		\hline
		 &  Min Groud Dis.(m). & Min Base Dis.(m)  \\ 
		\hline
		Compliance & $0.46 \pm 0.20$ &  $0.29 \pm 0.12$ \\
		\hline
	\end{tabular}
	\label{table4}
\end{table}

\begin{table}[t]
	\centering
	\caption{Similarity statistics of trajectories}
	\renewcommand{\arraystretch}{1.5} % 设置表格行高为原始行高的1.5倍
	\setlength{\tabcolsep}{4pt} 	  % 设置列之间的间隔为3pt(缩小间隔)
	\begin{tabular}{cccc}
		\hline
		 &  Intra-group Dis.(m) & Inter-group Dis.(m) &  P Value  \\ 
		\hline
		DTW Distance & $8.39\pm 4.35$ & $9.12\pm4.72$ &  0.61 \\
		\hline
	\end{tabular}
	\label{table5}
\end{table}

\begin{table}[t]
	\centering
	\caption{Comparison of peak torque}
	\renewcommand{\arraystretch}{1.5} % 设置表格行高为原始行高的1.5倍
	\setlength{\tabcolsep}{4pt} 	  % 设置列之间的间隔为3pt(缩小间隔)
	\begin{tabular}{cccc}
		\hline
		 &  Primeval & Compliance &  P Value \\ 
		\hline
		Max Peak Torque &  $0.87\pm0.11$ & $0.62\pm0.08$ &  $1.17e-09$    \\
		\hline
	\end{tabular}
	\label{table6}
\end{table}

\subsection{Real-World Scenarios}
As shown in Fig. \ref{physical}, a real-world experiment is conducted on a mobile manipulator. We use a C-100 differential mobile base ($n_b$ = 2) and a franka panda manipulator ($n_m$ = 7), interfacing with the robot through ROS Noetic. NOKOV motion capture provides position and velocity measurements, which are used for both ball state estimation and mobile base state tracking. We use a standard tennis ball with a reflective strip (diameter: 6.8 cm, mass: 60.3 g) for tracking purposes, along with an inelastic mesh bag as the container which would not weaken the impact. Due to the coupling effects of differential wheel steering, our high-level planner considers only the combined motion $q_1$ and restricts the differential motion $q_0$ to 0 to achieve more accurate tracking in the PRC. When a human hand throws the ball, a Kalman filter models the ball’s dynamics using the initial frames and updates the model in the subsequent few frames. This model is then used to predict the ball’s trajectory over the next 1.5s. After that, the High-Level planner, PRE planner, and network inference will be executed sequentially, and each will only be executed once. It is worth noting that we tested the delay in ROS master-slave communication and the robot's velocity control startup time. This delay was accounted for by sending the POC planning command in advance, allowing the robot to start the POC planning right around the time of the catch.

\begin{figure*}[t]
	\centering
	\includegraphics[width= \textwidth,height=4cm]{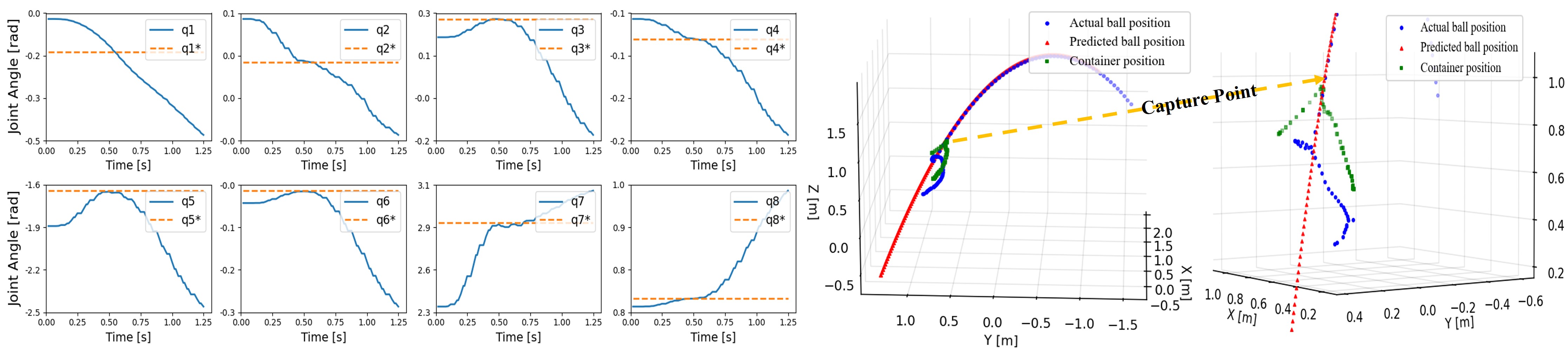}
	\caption{The change of the joint angles of the robot during a catch, the expected joint angles calculated by the high-level planner, and the trajectory of the ball and the container.}
\label{qt}
\end{figure*}

\begin{figure}[t]
	\centering
	\includegraphics[width=0.44\textwidth ,height=3.5cm]{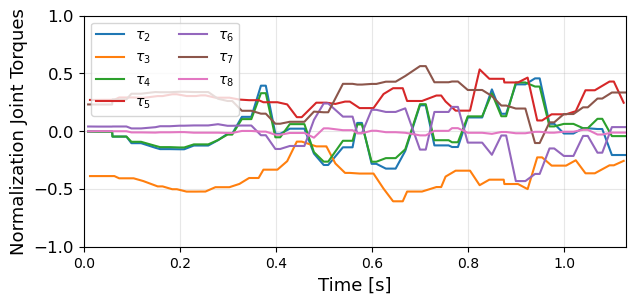}
	\caption{The curves of normalized joint torques. The maximum peak torque of all joints is 0.606. }
\label{joint torque}
\end{figure} 

We conducted a total of 100 hand-throwing experiments for compliant catching. We assess the effectiveness of the hand-thrown ball based on the closest distance between the container's initial position and the ball's trajectory. Due to inaccuracies in hand throwing, 19 attempts were deemed invalid. Out of the remaining 81 valid throws, the robot successfully captured the ball 75 times, resulting in a success rate of 92.59\%, without any collision. The failure occurred because noise in the capture system caused a discrepancy between the filter's estimated state and the actual state, leading to deviations in the predicted trajectory. It is worth mentioning that, due to the limitation and some obstacles of the laboratory room (some areas wound do not capture the ball well), we did not test in all directions like the simulation, but we still tried to throw to different directions of the robot (left, right, top and middle). In all experiments, TABLE \ref{table3} presents the error between the container's capture point and the actual ball position ``cap. Error", along with the capture time ``cap. Time" and the total calculation time ``cal. Time" for the Hive-Level, the PRC, and the Network inference phase. TABLE \ref{table4} provides collision avoidance data, including the minimum distances between the container and the ground ``Min Ground Dis.", as well as the container and the mobile base ``Min Base Dis.'', during the POC phase. We successfully achieved a high precision capture and collision avoidance.

We compare the effect of adding compliance to reduce torque impact. Since different ball trajectories can result in different torque impacts, ensuring trajectory similarity in the comparison is crucial. To control for variables, we ask the person to consistently face the same direction to maintain similarity and record data 30 times for both the compliant and non-compliant conditions. In each of the 30 trajectories, we quickly select the 20 most similar ones  by calculating the vector from the container's initial position to the closest point on trajectory. We then use Dynamic Time Warping (DTW) to assess the similarity between the different group trajectories. The intra-group distance for the compliant group and the inter-group distance between the compliant and non-compliant groups have been calculated, as shown in TABLE \ref{table5}.  The final p-value is much greater than the significance level of 0.05, indicating that the ball throwing trajectories of the two groups are similar. Therefore, the impact analysis can proceed to the next step. 

Due to different hardware in each joint, every torque was normalized according to the maximum allowed torque at that joint, thus, ${\tau'}_{i}=\tau_{i} / \tau_{i}^{\max }, i \in[2,8]$. We calculate the maximum peak torque $max(\tau'_i)$ in all joints during the experiment, as the ``Max Peak Torque" in TABLE \ref{table6}. The p-value for the ``Max Peak Torque" between the two groups is significantly lower than the 0.05 significance level, indicating a notable difference between the groups and confirming the feasibility of our compliant strategy. Using the mean value calculation, the quantitative attenuation ratio is $(0.87-0.62)/0.87 = 28.7 \% $. Since there is currently no research specifically on compliant catching for mobile manipulators, we referred to the compliant ratio data for a fixed manipulator catching a vertically falling object \cite{Zhao2023b}, which is 28\%. This result is similar to ours. However, our study is conducted in a more complex catching scenario and with a mobile manipulator. For one test in compliant catching, Figure \ref{qt}(a) shows the expected output of the high-level capture planner, along with the changes in joint angles during the catching process. At the same time, Figure \ref{qt}(b) presents the actual predicted trajectory of the ball and the container's trajectory during the compliant catching experiment. Additionally, Fig. \ref{joint torque} illustrates the normalized joint torques.

\section{CONCLUSIONS}
This paper aims to address the challenge of compliant catching for mobile manipulators. We have developed a novel CCMM framework that combines optimization and learning to address all the planning and tracking challenges during the catching process. A high-level planner is used to calculate the optimal capture point and joint configuration, while a PRC planner ensures the manipulator reaches the target joint configuration as quickly as possible. The compliant strategies are learned from human demonstrations, with a P-LSTM network proposed to capture temporal dependencies and spatial context through positional encoding. Additionally, a POC planner is proposed to track the sequence predicted by the network while addressing safety concerns arising from structural differences between humans and robots. Experimental results show a 98.70\% success rate in simulations, 92.59\% in real-world tests, and a 28.7\% reduction in impact torque.
The limitation of our framework is its reliance on the accuracy of external sensing, as the data from the initial frames is crucial for subsequent predictions and planning. Moving forward, a key focus will be exploring how to complete above tasks using only the robot's camera. Finally, due to the modular design of our CCMM framework, we envision that it would be easily ported to other robots or fields.

\bibliographystyle{IEEEtran}
\bibliography{ref}

\addtolength{\textheight}{-12cm}   % This command serves to balance the column lengths
                                  % on the last page of the document manually. It shortens
                                  % the height of the last page by a suitable amount.
                                  % This command does not take effect until the next page
                                  % so it should come on the page before the last. Make
                                  % sure that you do not shorten the height too much.

\end{document}